\documentclass[runningheads]{llncs}
\usepackage{graphicx}
\usepackage{multirow}
\usepackage{setspace}
\usepackage{caption}
\usepackage{amsfonts}
\usepackage{amsmath}
\usepackage{subfiles}
\usepackage{subcaption}
\usepackage[hidelinks,urlcolor=blue]{hyperref}
\usepackage[symbol]{footmisc}

\newcounter{daggerfootnote}

\begin{document}
\title{Goal-conditioned reinforcement learning for ultrasound navigation guidance}
%

\author{
Abdoul Aziz Amadou\inst{1,2} \and
Vivek Singh \inst{3} \and
Florin C. Ghesu \inst{4} \and
Young-Ho Kim \inst{3} \and
Laura Stanciulescu \inst{3,5} \and
Harshitha P. Sai \inst{6,7} \and
Puneet Sharma \inst{3} \and
Alistair Young \inst{1} \and
Ronak Rajani \inst{1,8} \and
Kawal Rhode \inst{1}
}

\institute{King’s College London, School of Biomedical Engineering \& Imaging Sciences, London, United Kingdom \\
\email{abdoul.a.amadou@kcl.ac.uk} \\ \and
Siemens Healthcare Limited, Camberley, United Kingdom\\ \and
Siemens Healthineers, Digital Technology and Innovation, Princeton, NJ, USA \\ \and
Siemens Healthineers AG, Digital Technology and Innovation, Erlangen, Germany \\ \and
Carol Davila University of Medicine and Pharmacy Bucharest, Romania \\ \and
Siemens Healthineers, Sonographer , Bangalore, Karnataka \and
Fortis Institute of Medical Sciences, Affiliated by Rajiv Gandhi University of Medical Sciences, Department of Cardiology, Bangalore, Karnataka \and
Guy’s and St Thomas’ NHS Foundation Trust, London, United Kingdom\\
}
\authorrunning{A.A Amadou et al.}

%

%
%
\maketitle              
\begin{abstract}
Transesophageal echocardiography (TEE) plays a pivotal role in cardiology for diagnostic and interventional procedures. However, using it effectively requires extensive training due to the intricate nature of image acquisition and interpretation. To enhance the efficiency of novice sonographers and reduce variability in scan acquisitions, we propose a novel ultrasound (US) navigation assistance method based on contrastive learning as goal-conditioned reinforcement learning (GCRL). We augment the previous framework using a novel contrastive patient batching method (CPB) and a data-augmented contrastive loss, both of which we demonstrate are essential to ensure generalization to anatomical variations across patients. The proposed framework enables navigation to both standard diagnostic as well as intricate interventional views with a single model. Our method was developed with a large dataset of 789 patients and obtained an average error of 6.56 mm in position and 9.36 degrees in angle on a testing dataset of 140 patients, which is competitive or superior to models trained on individual views. Furthermore, we quantitatively validate our method's ability to navigate to interventional views such as the Left Atrial Appendage (LAA) view used in LAA closure. Our approach holds promise in providing valuable guidance during transesophageal ultrasound examinations, contributing to the advancement of skill acquisition for cardiac ultrasound practitioners.

\end{abstract}

\section{Introduction}

Echocardiography is a key imaging modality in the diagnosis and treatment of cardiovascular diseases. While several US modalities are used in practice, in TEE, the transducer images the heart from the oesophagus, often yielding better scan quality and helping circumvent issues caused by acoustic windows in other modalities such as transthoracic echocardiography (TTE). Training operators for TEE is time-consuming due to complex controls and image interpretation, with an added risk of patient injury due to incorrect transducer manipulation. Additionally, in structural heart procedures where TEE is coupled with fluoroscopy, health issues arise for catheterization lab staff due to orthopaedic strain and radiation exposure \cite{Andreassi2016OccupationalHR}.

AI-assisted guidance for transducer manipulation has been proven to benefit operator training, lower the learning curve, and reduce intra and inter-user variability \cite{Narang2021UtilityOA,sabo2023}. Additional advantages include shortening of TEE examinations, enhancing patient comfort and reducing radiation exposure during interventional procedures.

Various deep reinforcement learning (DRL) approaches for ultrasound autonomous navigation have been proposed, primarily focusing on extracorporeal scanning of anatomies like the spine \cite{Li2021AutonomousNO,Hase2020UltrasoundGuidedRN} and neck \cite{Bi2022VesNetRLSR}. However, Li et al. \cite{Li2021AutonomousNO} suffer from a lack of generalization to unseen patient datasets, and \cite{Hase2020UltrasoundGuidedRN} employs simplified state and action spaces that do not capture the real-world scanning conditions well. While previous works rely on simulation environments, both \cite{Droste2020AutomaticPM,milletari_straight_2019} use additional hardware attached to the transducers to acquire datasets for imitation learning. However, the scalability of the data acquisition (time and cost) is reported as one of the limitations of such approaches \cite{milletari_straight_2019}. TEE imaging has been less explored, with Wang et al. \cite{Wang2021RoboticIU} using a simulation environment based on segmented pre-operative scans to find robotic poses corresponding to desired views pre-operatively. However, this approach requires a manual intervention to define the views. Finally, authors in Li et al. \cite{Li2023RLTEEAP} use a simulation environment to train models to navigate to standard TEE views. However, their approach involves training one model for each target view, which does not scale well to support additional views or supporting manoeuvres to visualize specific structures more clearly. Furthermore, they only control 3 out of 5 transducer degrees of freedom and test on a limited dataset of 5 patients.

\begin{figure}[t!]
\centering
\includegraphics[width=0.95\textwidth]{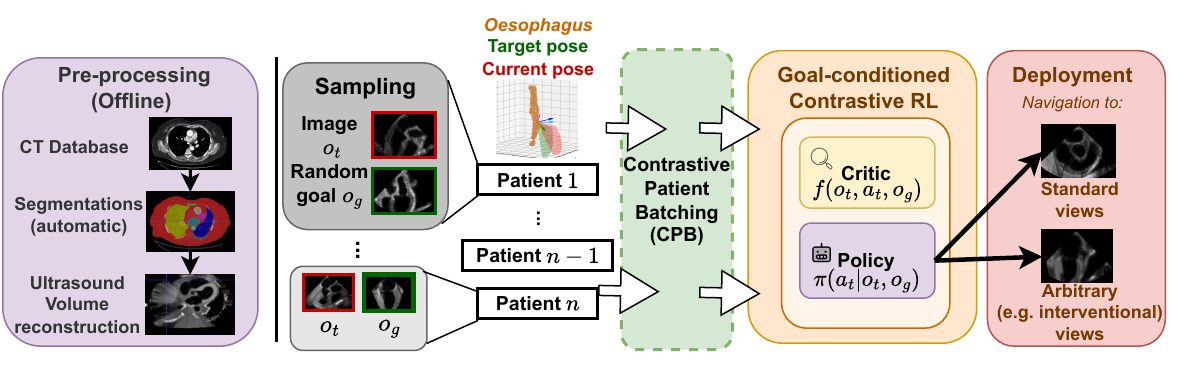}
\caption{\textit{System overview of Goal-conditioned RL for Ultrasound Navigation}. We first segment CTs and generate ultrasound volume reconstructions for rapid sampling during training. The model is trained to reach randomly selected goal views by employing the contrastive patient batching (CPB) mechanism to create a contrastive batch from the collected experience. When deployed, the trained model can navigate to arbitrary views, including standard and interventional views.} \label{fig:pipeline}
\end{figure}

This paper introduces a novel approach to training a navigation model using goal-conditioned reinforcement learning. We build upon Contrastive RL (CRL) \cite{Eysenbach2022ContrastiveLA}, a state-of-the-art goal-conditioned method which showed promising results in image-based robotic tasks. We train our model using random goal views, enabling navigation to arbitrary views given a user-defined goal. We make use of a simulation environment \cite{amadou2024cardiac}, where we leverage a large dataset of chest and cardiac CTs to train our model and enable generalization to unseen patients. An overview of the proposed workflow is shown in Fig. \ref{fig:pipeline}.

The contributions of this work are the following:
\textbf{1)} We propose a novel methodology for TEE imaging guidance to arbitrary views using goal-conditioned reinforcement learning. This not only enables navigation to standard views but also to alternative views showing specific structures. \textbf{2)} We enable the generalization of the CRL framework by introducing: \textit{(i)} Contrastive patient batching, a simple yet effective method to sample hard contrastive pairs and improve performance; \textit{(ii)} A novel contrastive data augmentation loss to improve both robustness and the quality of learnt representations. \textbf{3)} We demonstrate the effectiveness of our approach by performing two experiments on a dataset of 140 patients: \textit{(i)} By navigating to standard views, including views that were not explicitly sampled during training. Our method achieves competitive performance to RL methods trained to reach individual views; \textit{(ii)} By navigating to a non-standard view used to monitor the deployment of devices in LAA closure. This showcases the usability of our method both for diagnostic and interventional cases. To the best of our knowledge, this work is the first attempt to develop an ultrasound navigation model capable of navigating to arbitrary views given a goal.

\section{Methodology}

\subsection{Simulation environment}

Acquiring real datasets for navigation is a cumbersome, expensive and time-consuming task, as reported in \cite{Droste2020AutomaticPM,milletari_straight_2019}. Hence, as shown in Fig. \ref{fig:pipeline}, we employ a physics-based Computed Tomography (CT) to ultrasound simulation pipeline to train our model. \cite{amadou2024cardiac}. The pipeline takes as input chest and cardiac CTs and automatically segments them to obtain masks of the organs of interest, namely the oesophagus, heart chambers, aorta, lungs and pulmonary artery. A Monte Carlo path tracing algorithm is then used to simulate ultrasound wave propagation in tissue. The pipeline was extensively validated with phantom experiments, where US image properties were assessed, and a view classification experiment, in which we demonstrated the usefulness of the pipeline in generating data for model training. More details on the pre-processing and US simulation pipeline are provided in the anonymized submission in the supplementary material.

As simulating ultrasound images on the fly from the CT is computationally expensive and would significantly slow down the training, we followed Li et al. \cite{Li2023RLTEEAP} and generated simulated US images by translating the transducer down the oesophagus, rotating it by 360 degrees at every position. Simulation and volume reconstruction were done offline on the GPU. 

\begin{figure}[t!]
\centering
\includegraphics[width=\textwidth]{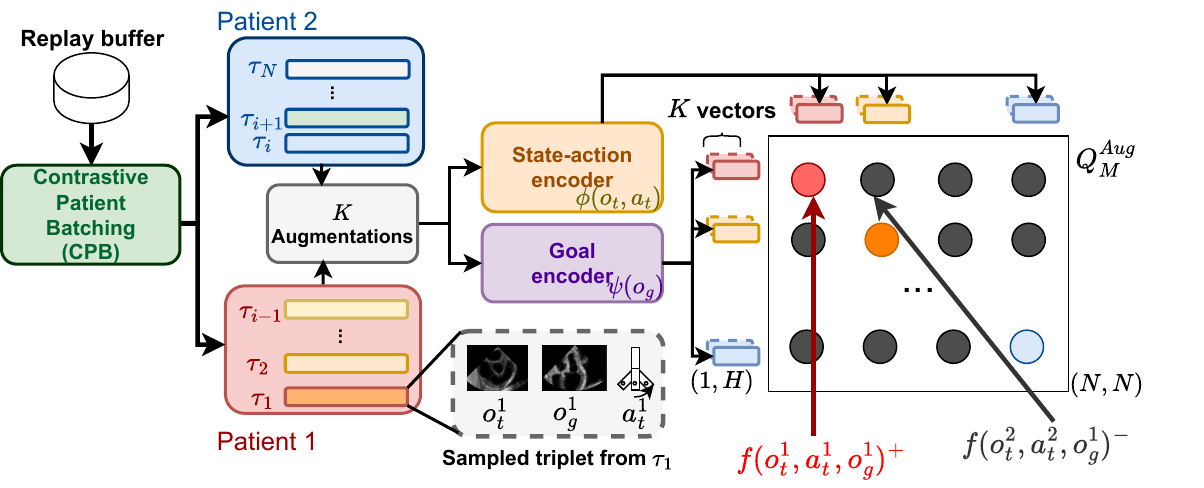}
\caption{\textit{Contrastive critic training:} We build a contrastive batch using trajectories from two patients with CPB and pass (observation, action) pairs and goal images to the state-action and goal encoders respectively. Goal and observations are augmented $K$ times and we build $K^2$ intermediate matrices (not shown) from the inner product between all the encoded representations, with $Q_M^{Aug}$ as their average. The critic is trained to maximize the similarity between state-action and goal representations of the same trajectories, which corresponds to the diagonal of the matrices.} \label{fig:losses}
\end{figure}

\subsection{Goal-Conditioned Reinforcement Learning}

The goal-conditioned navigation task is defined by: $S$, which represents the environment's state, defined by the transducer's pose in the CT coordinate system; $A$ is the set of TEE transducer movements, i.e. translation along the oesophagus, transducer rotation, the electronic rotation of the scanning plane and the left/right and retro/ante flexions; $p(s_{t+1}|s_t, a_t)$ are the transition probabilities between $s_{t+1}$ and $s_t$ after taking action $a_t$; $r_g(s, a) = (1 - \gamma)p(s_{t+1} = s_g | s_t, a_t)$ is the goal-conditioned reward function, defined as the probability density of reaching the goal $s_g$ at the next step; $\Omega$ is the set of observations $o$, which correspond to ultrasound images acquired from the transducer in a given state $s$; $\gamma \in [0, 1]$ the discount factor.

Our goal-conditioned framework follows the actor-critic architecture, where the critic takes as input a triplet of (observation, action, goal) $(o_t, a_t, o_g)$ and returns the probability (density) of reaching goal $o_g$ when taking action $a_t$ when given an observation $o_t$. The actor takes as input a pair $(o_t, o_g)$ and returns the action $a_t$ to take to reach the goal. The critic is trained to correctly predict which actions lead to a goal, and the actor learns to output correct actions by maximizing the critic's output.

Similarly to \cite{Eysenbach2022ContrastiveLA}, contrastive learning is used to train a critic function by making use of two models $\phi$ and $\psi$ which encode state-action (SA) pairs  $(o_t, a_t)$ and goals $o_g$ respectively. The critic function measures the similarity of the latent representations of dimension $H$ via inner-product $f(o_t, a_t, o_g) = \langle \phi(o_t, a_t), \psi(o_g) \rangle$, as illustrated in Fig. \ref{fig:losses}. When an action likely leads to a goal from a given pose, the inner product will have a high value, indicating the probability (density) of reaching the goal is high.

During training, the transducer is initialized at a given pose $s_0$ (yielding observation $o_0$) and is given a goal observation $o_g$. The sequence of observations/actions until the last timestep gives a trajectory $\tau_i = (o_0^i, a_0^i, o_1^i, ..., o_n^i)$.

\textbf{Critic loss:} To train the critic, as illustrated in Fig. \ref{fig:losses}, we sampled the input triplet from a trajectory $(o_t^i, a_t^i, o_g^i) \sim \tau_i$, where the positive goal timestep $T$ is a future timestep ($T > t$) sampled from a geometric distribution $T \sim Geom(1 - \gamma)$. The negative goal $o_{g'}^j$ is sampled randomly from another trajectory $\tau_j$. The critic loss is based on the infoNCE loss \cite{vanderoord2018} and computed as:

\begin{equation}
    \max_{f} \mathbb{E}_{\substack{(o_t^i, a_t^i, o_g^i) \sim \tau_i \\ o_{g'} \sim \tau_j }} log\big[\frac{e^{f(o_t^i, a_t^i, o_g^i)^+}}{e^{f(o_t^i, a_t^i, o_g^i)^+} + \sum_j e^{f(o_t^i, a_t^i, o_{g'}^j)^-}}\big]
    \label{eq:critic_loss}
\end{equation}

Where $f(o_t, a_t, o_g)^+$ and, $f(o_t, a_t, o_g)^-$ denote the critic output for positive and negative examples respectively. The inner product between all SA and goal representations in a batch of size $N$ gives a matrix $Q_M$ of size $(N, N)$ on which we apply a cross-entropy loss row and column-wise, with the true labels being on the diagonal. In order to stabilize the critic training, we observed that the normalization of the goal representations was necessary. Furthermore, the use of a temperature scaling parameter of state-action representations, combined with L2 regularization was necessary. Hyperparameters are listed in the supplementary material.

\textbf{Actor loss:} The actor takes as input observation and goal pairs $(o_t, o_g)$ and returns an action $a_t$ to reach the goal. The actor simply aims at maximizing the critic output such that:

\begin{equation}
    \max_{\pi} \mathbb{E}_{\substack{ a_t \sim \pi(o_t, o_g)}} f(o_t, a_t, o_g)
    \label{eq:actor_loss}
\end{equation}

In practice, the actor outputs the mean and standard deviation of a multivariate Gaussian from which we sample and apply tanh squashing to obtain the bounded actions.
During training, we noticed that using random goals sampled from other trajectories rather than goals from the same trajectory as $o_t$ led to better actor performance, as also reported in \cite{Eysenbach2022ContrastiveLA}.

\textbf{Data augmented contrastive loss:} Given a triplet $(o_t, a_t, o_g)$, the critic output should be similar to the data augmented triplet $(o'_t, a_t, o'_g)$, where $o'_t$ is a randomly shifted version of $o_t$. We apply $K$ random shifts to the observations and goal images, where the k-th augmentation is denoted as $o_{t,k}$. The critic loss is then computed on the average of the $K^2$ matrices $Q_M^i$ resulting from the inner products of the augmented observations and goals.
\begin{equation}
    Q_M^{aug} = \mathbb{E}_i[Q^i_M] = \frac{1}{K^2} \sum_{k=1}^K \sum_{k'=1}^K f(o_{t,k}, a_t, o_{g,k'})
    \label{eq:data_aug_loss}
\end{equation}

\textbf{Contrastive Patient Batching (CPB):} We observed empirically that the composition of the contrastive batch plays a significant role in the convergence of the critic. Following the strategy proposed in \cite{Eysenbach2022ContrastiveLA}, where samples are randomly chosen from the replay buffer yields poor results in our setting. A closer investigation revealed that a randomly sampled batch contains samples from different patients with different intermediate states, and the critic ends up learning features associated with anatomical differences between patients, rather than general anatomical features necessary for the control task. To address this, we tag the trajectories by the corresponding patient identifier during training. While creating a  batch of size $N$, we sample $(o_t, a_t, o_g)$ triplets from two patients, with $\frac{N}{2}$ samples per patient. Having a significant number of samples coming from the same patient creates harder negatives for the critic, which improves its effectiveness in discriminating trajectories. Ablation studies in the supplementary material show performance for different numbers of patients per batch.

\begin{table}[t!]
\centering
\caption{Quantitative results (mean $\pm$ std) for the standard view navigation experiment. Goal type \textit{Patient/Template} indicates whether the input goal was generated from the same patient or from a template patient. Note that no perturbations were explicitly sampled around the ME 5CH view during training. (*) Results for RL-TEE and SAC are obtained from several models, each one trained separately on a view. CRL+B indicates CRL-D trained with CPB and CRL + BA is CRL+B with the data augmented contrastive loss.}
\begin{tabular}{|c|c|c|c|c|}  
 \hline
 Views & Goal type & Method & Angle Error (deg) & Position error (mm) \\
\hline
\hline
 \multirow{5}{1.8cm}{ME AV SAX, 2CH, 4CH, LAX} & \multirow{2}{*}{N/A} & RL-TEE \cite{Li2023RLTEEAP}* & 9.90 $\pm$ 8.04 & 9.17 $\pm$ 6.87  \\
 \cline{3-5}
& & SAC* \cite{Haarnoja2018SoftAO} & 9.77 $\pm$ 10.89 & 7.92 $\pm$ 9.35 \\
\cline{2-5}
& \multirow{3}{*}{Patient} & CRL-D \cite{Eysenbach2022ContrastiveLA} & 18.47 $\pm$ 19.89 & 13.27 $\pm$ 17.96 \\
\cline{3-5}
& & CRL+B  & \textbf{9.00 $\pm$ 12.37} & \textbf{5.93 $\pm$ 8.17}\\
\cline{3-5}
& & CRL+BA & 9.36 $\pm$ 9.52 & 6.56 $\pm$ 6.46\\
\hline
\hline
\multirow{3}{*}{ME 5CH} & \multirow{3}{*}{Patient} & CRL-D \cite{Eysenbach2022ContrastiveLA}  & 35.24 $\pm$ 25.67 & 15.23 $\pm$ 15.08 \\
\cline{3-5}
& & CRL+B & 19.88 $\pm$ 35.91 & 8.88 $\pm$ 11.02 \\
\cline{3-5}
& & CRL+BA & \textbf{11.40 $\pm$ 5.30} & \textbf{7.80 $\pm$ 4.11} \\
\hline
\hline
\multirow{3}{1.5cm}{ME 2CH,4CH} & \multirow{3}{*}{Template} & CRL-D \cite{Eysenbach2022ContrastiveLA} & 28.91 $\pm$ 21.86 & 24.59 $\pm$ 24.63\\
\cline{3-5}
& & CRL+B & 13.39 $\pm$ 7.74 & 12.95 $\pm$ 7.32\\
\cline{3-5}
& & CRL+BA & \textbf{12.19 $\pm$ 6.88} & \textbf{10.54 $\pm$ 6.34} \\
\hline
\end{tabular}
\label{table:standard_view_experiment}
\end{table}

\textbf{Training loop:} We automatically find probe poses to obtain standard views using landmarks extracted from the automatic segmentations and by following clinical guidelines \cite{hanh2013}. When applying actions, the transducer is translated along the oesophagus centerline and we constrain all motions to remain within its walls. At the start of each episode, we initialize the transducer at one of the standard view poses. Random perturbations are applied to obtain the starting pose $s_0$. For CRL, we obtain a goal pose by applying additional random perturbations from $s_0$, yielding a goal pose $s_g$. Hence our model is always trained with random goals and never explicitly trained to navigate to a standard view. Perturbation ranges are listed in the supplementary material.

\section{Experiments and results}
\textbf{Standard view navigation: } We first compare our approach with existing methods by examining their ability to reach standard views, a task essential in TEE examinations. We processed 929 patient CT datasets from the LIDC-IDRI dataset \cite{lidc_idri}: 653 were used for training, 136 for validation and the rest 140 for testing. Goal images in the test dataset were reviewed and confirmed by a cardiologist. We evaluated our method in two scenarios based on how the goal is specified: Using the (synthetic) US view generated from the same patient as the goal or using a US view from another "template" patient as a goal. The latter corresponds to a scenario where a user may not have access to prior scans of the patient and hence uses a similar view from another reference to specify the goal. 
We report the position and angle error at the end of the episode w.r.t the ground truth pose. For each patient and goal pair, we ran 10 experiments with the transducer initialized at random positions. Results are reported in Table \ref{table:standard_view_experiment} and a breakdown of the results per view is included in the supplementary material, alongside videos showing the navigation process.

We compare the performance of our model with  Li et al. \cite{Li2023RLTEEAP} (RL-TEE), Soft Actor-Critic (SAC) \cite{Haarnoja2018SoftAO} which is a state-of-the-art off-policy reinforcement learning algorithm and the default implementation of CRL (CRL-D)\cite{Eysenbach2022ContrastiveLA}. For RL-TEE and SAC, we train one model per view as the algorithms are not goal-conditioned. SAC models are trained using the same rewards as in \cite{Li2023RLTEEAP}. We use four mid-oesophageal (ME) views for training: Two and four chambers (2CH, 4CH), long-axis (LAX) and aortic valve short-axis (AV SAX). Due to the similarity between ME AV SAX and ME Right Ventricle Inflow-Outflow views, samples resembling one or the other class were considered to be of the AV SAX class. We use templates from ME 2CH and 4CH views as they are better geometrically defined across patients.

Finally, we showcase the versatility of our method by inputting ME five chambers (5CH) views as goals to the model during testing. ME 5CH views were not used as a starting point for random perturbations during training. In Table \ref{table:standard_view_experiment}, CRL+B model corresponds to CRL + CPB, and CRL+BA is CRL + CPB + data augmented contrastive loss.

\begin{figure}[h!]
\centering
\includegraphics[width=0.9\textwidth]{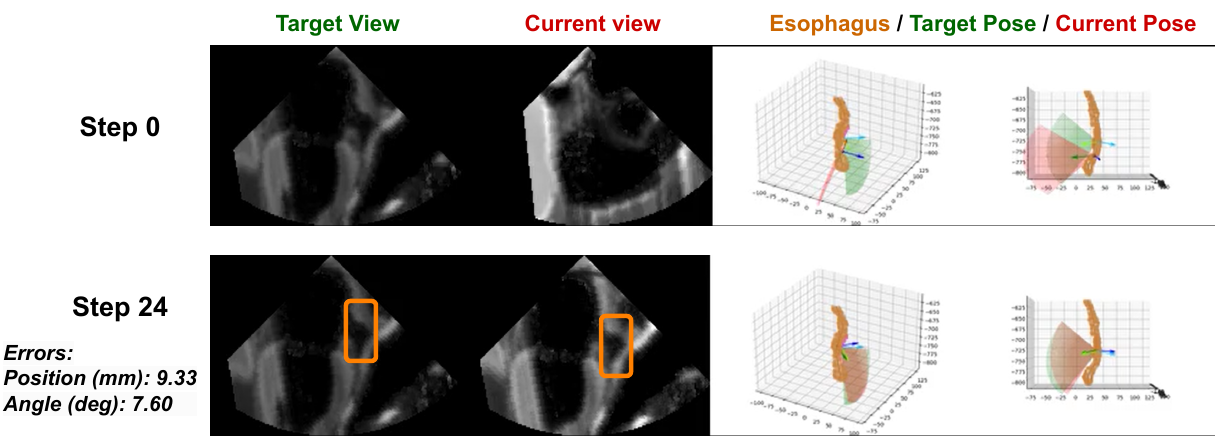}
\caption{Example navigation to a view showing the LAA (orange box). The two rightmost pictures are projections showing the desired (green) and current (red) transducer positions.}
\label{fig:navigation_example}
\end{figure}

\textbf{Interventional view navigation:} In a second experiment, we showcase the usefulness of goal-conditioning by navigating to a non-standard view used in LAA closure procedures. We use the FUMPE dataset \cite{fumpe} (train: 21 / test: 5) for which we have additional LAA segmentations, hence finetuning is required as the LAA was missing in the previous datasets. Previously trained CRL models are finetuned for 250K steps, without changing the training procedure or sampling trajectories near the LAA explicitly. Quantitative results are reported in Table \ref{table:laa_experiment}, where the performance is on par with standard view navigation. Qualitative results are shown in Fig. \ref{fig:navigation_example}.\\
For all experiments and models, we use a ResNet-18 \cite{he2015} as image an encoder, Adam optimizer \cite{Kingma2014AdamAM} and train with A4500 GPUs. Detailed result tables and demonstration videos are included in the supplementary material.

\begin{table}[t!]
\centering
\caption{Quantitative results for the LAA view navigation experiment. The high-quality representations learnt by the model with the data augmented contrastive loss allow for better generalization and transfer.}
\begin{tabular}{|c|c|c|c|c|}
\hline
Views & Goal type & Method & Angle Error (deg) & Position error (mm) \\
\hline
\hline
 \multirow{3}{*}{LAA} & \multirow{3}{*}{Patient} & CRL-D \cite{Eysenbach2022ContrastiveLA} & 37.70 $\pm$ 28.74 & 30.99 $\pm$ 28.61\\
\cline{3-3}\cline{4-5}
& &  CRL+B & 24.63 $\pm$ 23.69 & 17.54 $\pm$ 17.22\\
\cline{3-3}\cline{4-5}
& &  CRL+BA & \textbf{10.18 $\pm$ 5.58} & \textbf{9.02 $\pm$ 4.33}\\
\hline
\end{tabular}
\label{table:laa_experiment}
\end{table}

\section{Discussion and conclusion}

\textbf{Discussion:} Our generalist model achieves competitive performance to specialist models trained to navigate to single views, whether it is given goal images from the same or a template patient. Additionally, the model robustly navigates to arbitrary views without explicit sampling during training, as shown by the results on ME 5CH and LAA views, thus demonstrating the versatility of the goal-conditioned framework. Note that the performance in such scenarios is highly dependent on the agent's exploration of the environment during training. A drawback of CRL is the longer training time, as the contrastive critic needs many samples to converge. We alleviate this with an efficient asynchronous implementation using RLLib \cite{ray2017}, yielding a training time of two days for 200M steps. Finally, deployment in a real-world setting would potentially require fine-tuning using either real data and/or improved simulations with generative models to address any reality gap.\\
\textbf{Conclusion:} We have presented a novel approach for ultrasound navigation using goal-conditioned reinforcement learning. Given a goal image, our versatile model navigates robustly both to standard and arbitrary views showing specific structures. Using this method as a guidance system could help train sonographers, improve the acquisition quality and reduce variability among experienced users.

\noindent \textbf{Acknowledgements.} The authors acknowledge the National Cancer Institute and the Foundation for the National Institutes of Health, and their critical role in the creation of the free publicly available LIDC/IDRI Database used in this study.
For the purpose of open access, the author has applied a CC BY public copyright licence to any Author Accepted Manuscript version arising from this submission

\noindent \textbf{Disclaimer.} The concepts and information presented in this paper are based on research results that are not commercially available. Future commercial availability cannot be guaranteed.\\

%
%
%
\newpage
\bibliographystyle{splncs04}
\bibliography{bibliography}

\end{document}